\documentclass[11pt,a4paper]{article}
\usepackage[hyperref]{acl2019}
\usepackage{times}
\usepackage{latexsym}
\usepackage{url}
\usepackage{amsmath}
\usepackage{tabularx}
\usepackage{booktabs}
\usepackage{amssymb}
\usepackage{bm}
\usepackage{multirow}
\usepackage{tcolorbox}
\usepackage[framemethod=tikz]{mdframed}
\usepackage{tikz,pgfplots}
\usepgfplotslibrary{groupplots}

\newcommand{\V}[1]{\bm{#1}}
\newcommand{\mscore}[1]{s_\text{m}(#1)}

\newcommand{\ffnn}[2]{\textsc{ffnn}_\text{#1}(#2)}

\definecolor{g-red}{HTML}{DB4437}
\definecolor{g-blue}{HTML}{4285F4}
\definecolor{g-green}{HTML}{0F9D58}
\definecolor{g-yellow}{HTML}{F4B400}
\definecolor{g-orange}{HTML}{FF9800}
\definecolor{g-grey}{HTML}{9E9E9E}
\definecolor{shannon}{HTML}{304FFE}
\definecolor{uw}{RGB}{138,43,226}
\definecolor{stanford}{RGB}{255,69,0}
\definecolor{const}{RGB}{68, 110, 182}
\definecolor{head}{RGB}{246, 180, 32}
\definecolor{freq}{RGB}{0, 0, 0}

 \aclfinalcopy 

\title{CorefQA: Coreference Resolution as Query-based Span Prediction}

\author{Wei Wu$^\clubsuit$, Fei Wang$^\clubsuit$, 
Arianna Yuan$^{\blacklozenge\clubsuit}$,
Fei Wu$^{\spadesuit}$ and Jiwei Li$^{\spadesuit\clubsuit}$ \\
$^{\spadesuit}$ Department of Computer Science and Technology, Zhejiang University\\
$^\blacklozenge$ Computer Science Department, Stanford University\\
$^{\clubsuit}$ ShannonAI \\
xfyuan@stanford.edu, wufei@zju.edu.cn  \\
 {\{wei\_wu, fei\_wang,jiwei\_li\}}@shannonai.com
}

\date{}
\begin{document}
\maketitle
\begin{abstract}
In this paper, we present CorefQA, an accurate and extensible approach for the coreference resolution task.
We formulate the problem as a span prediction task, like in question answering:
A query is generated for each candidate mention using its surrounding context, and a span prediction module is employed to extract the text spans of the coreferences within the document using the generated query.
This formulation comes with the following key advantages: 
(1) The span prediction strategy provides the flexibility of retrieving mentions left out at the mention proposal stage;
(2) In the question answering framework, encoding the mention and its context explicitly in a query makes it possible to have a deep and thorough examination of cues embedded in the context of coreferent mentions; and
(3) A plethora of existing question answering datasets can be used for data augmentation to improve the model's generalization capability.
Experiments demonstrate significant performance boost over previous models, with 83.1 (+3.5) F1 score on the CoNLL-2012 benchmark and 87.5 (+2.5) F1 score on the GAP benchmark.
\footnote{\url{https://github.com/ShannonAI/CorefQA}}
\end{abstract}

\section{Introduction}

Recent coreference resolution systems \citep{DBLP:conf/emnlp/LeeHLZ17, DBLP:conf/naacl/LeeHZ18, DBLP:conf/acl/ZhangSYXR18, DBLP:conf/acl/KantorG19} consider all text spans in a document as potential mentions and learn to find an antecedent for each possible mention. There are two key issues with this paradigm, in terms of task formalization and the algorithm. 

At the task formalization level, mentions left out at the mention proposal stage can never be recovered since the downstream module only operates on the proposed mentions. Existing models often suffer from mention proposal \citep{DBLP:conf/acl/ZhangSYXR18}. The coreference datasets can only provide a weak signal for spans that correspond to entity mentions because singleton mentions are not explicitly labeled.
Due to the inferiority of the mention proposal model, it would be favorable if a coreference framework had a mechanism to retrieve left-out mentions. 

\begin{figure}[t!]
\centering
\setlength{\fboxsep}{0.9em}
\fbox{\parbox{0.9\linewidth}{
 \textbf{Original Passage}
 
 In addition , \textcolor{red}{many people} were poisoned when \textcolor{blue}{toxic gas} was released.  \textcolor{red}{They} were poisoned and did not know how to protect \textcolor{red}{themselves} against \textcolor{blue}{the poison}. 
 
\textbf{Converted Questions}

Q1: Who were poisoned when toxic gas was released?

A1: [\textcolor{red}{They}, \textcolor{red}{themselves}]

Q2: What was released when many people were poisoned? 

A2: [\textcolor{blue}{the poison}]

Q3: Who were poisoned and did not know how to protect themselves against the poison?

A3: [\textcolor{red}{many people}, \textcolor{red}{themselves}]

Q4: Whom did they not know how to protect against the poison?

A4: [\textcolor{red}{many people}, \textcolor{red}{They}]

Q5: They were poisoned and did not know how to protect themselves against what?

A5: [\textcolor{blue}{toxic gas}]
}}
\caption{An illustration of the paradigm shift from coreference resolution to query-based span prediction. Spans with the same color represent coreferent mentions. Note that we use a more direct strategy to generate the questions based on the mentions.}
\label{fig:intro-example}
\end{figure}

At the algorithm level, existing end-to-end methods \citep{DBLP:conf/emnlp/LeeHLZ17, DBLP:conf/naacl/LeeHZ18, DBLP:conf/acl/ZhangSYXR18} score each pair of mentions only based on mention representations from the output layer of a contextualization model. This means that the model lacks the connection between mentions and their contexts. Semantic matching operations between two mentions (and their contexts) are performed only at the output layer and are relatively superficial. Therefore it is hard for their models to capture all the lexical, semantic and syntactic cues in the context. 

To alleviate these issues, we propose CorefQA, a new approach that formulates the coreference resolution problem as a span prediction task, akin to the question answering setting.
A query is generated for each candidate mention using its surrounding context, and a span prediction module is further employed to extract the text spans of the coreferences within the document using the generated query. Some concrete examples are shown in Figure \ref{fig:intro-example}. \footnote{This is an illustration of the question formulation. The actual operation is described in Section \ref{ssec:question_answering}.}

This formulation provides benefits at both the task formulation level and the algorithm level. 
At the task formulation level, since left-out mentions can still be retrieved at the span prediction stage, the negative effect of undetected mentions is significantly alleviated. At the algorithm level, by generating a query for each candidate mention using its surrounding context, the CorefQA model explicitly considers the surrounding context of the target mentions, the influence of which will later be propagated to each input word using the self-attention mechanism. 
Additionally, unlike existing end-to-end methods \citep{DBLP:conf/emnlp/LeeHLZ17, DBLP:conf/naacl/LeeHZ18, DBLP:conf/acl/ZhangSYXR18}, where the interactions between two mentions are only superficially modeled at the output layer of contextualization, span prediction requires a more thorough and deeper examination of the lexical, semantic and syntactic cues within the context, which will potentially lead to better performance. 

Moreover, the proposed question answering formulation allows us to take advantage of existing question answering datasets. Coreference annotation is expensive, cumbersome and often requires linguistic expertise from annotators. Under the proposed formulation, the coreference resolution has the same format as the existing question answering datasets \citep{DBLP:conf/emnlp/RajpurkarZLL16, DBLP:conf/acl/RajpurkarJL18, DBLP:journals/corr/abs-1908-05803}.
Those datasets can thus readily be used for data augmentation. We show that pre-training on existing question answering datasets improves the model's generalization and transferability, leading to additional performance boost.

Experiments show that the proposed framework significantly outperforms previous models on two widely-used datasets. Specifically, we achieve new state-of-the-art scores of 83.1 (+3.5) on the CoNLL-2012 benchmark and 87.5 (+2.5) on the GAP benchmark. 

\section{Related Work}
\subsection{Coreference Resolution}
Coreference resolution is a fundamental problem in natural language processing and is considered as a good test of machine intelligence \citep{DBLP:journals/aim/MorgensternDO16}. Neural network models have shown promising results over the years. Earlier neural-based models \citep{DBLP:conf/naacl/WisemanRS16, DBLP:conf/acl/ClarkM15, DBLP:conf/acl/ClarkM16} rely on parsers and hand-engineered mention proposal algorithms. 
Recent work \citep{DBLP:conf/emnlp/LeeHLZ17, DBLP:conf/naacl/LeeHZ18, DBLP:conf/acl/KantorG19} tackled the problem in an end-to-end fashion by jointly detecting mentions and predicting coreferences. Based on how entity-level information is incorporated, they can be further categorized as (1) entity-level models \cite{DBLP:conf/acl/BjorkelundK14, DBLP:conf/acl/ClarkM15, DBLP:conf/acl/ClarkM16, DBLP:conf/naacl/WisemanRS16} that directly model the representation of real-world entities and (2) mention-ranking models \citep{DBLP:conf/emnlp/DurrettK13, DBLP:conf/acl/WisemanRSW15, DBLP:conf/emnlp/LeeHLZ17} that learn to select the antecedent of each anaphoric mention.
Our CorefQA model is essentially a mention-ranking model, but we identify coreference using question answering.

\begin{figure*}
\centering
\includegraphics[scale=0.483]{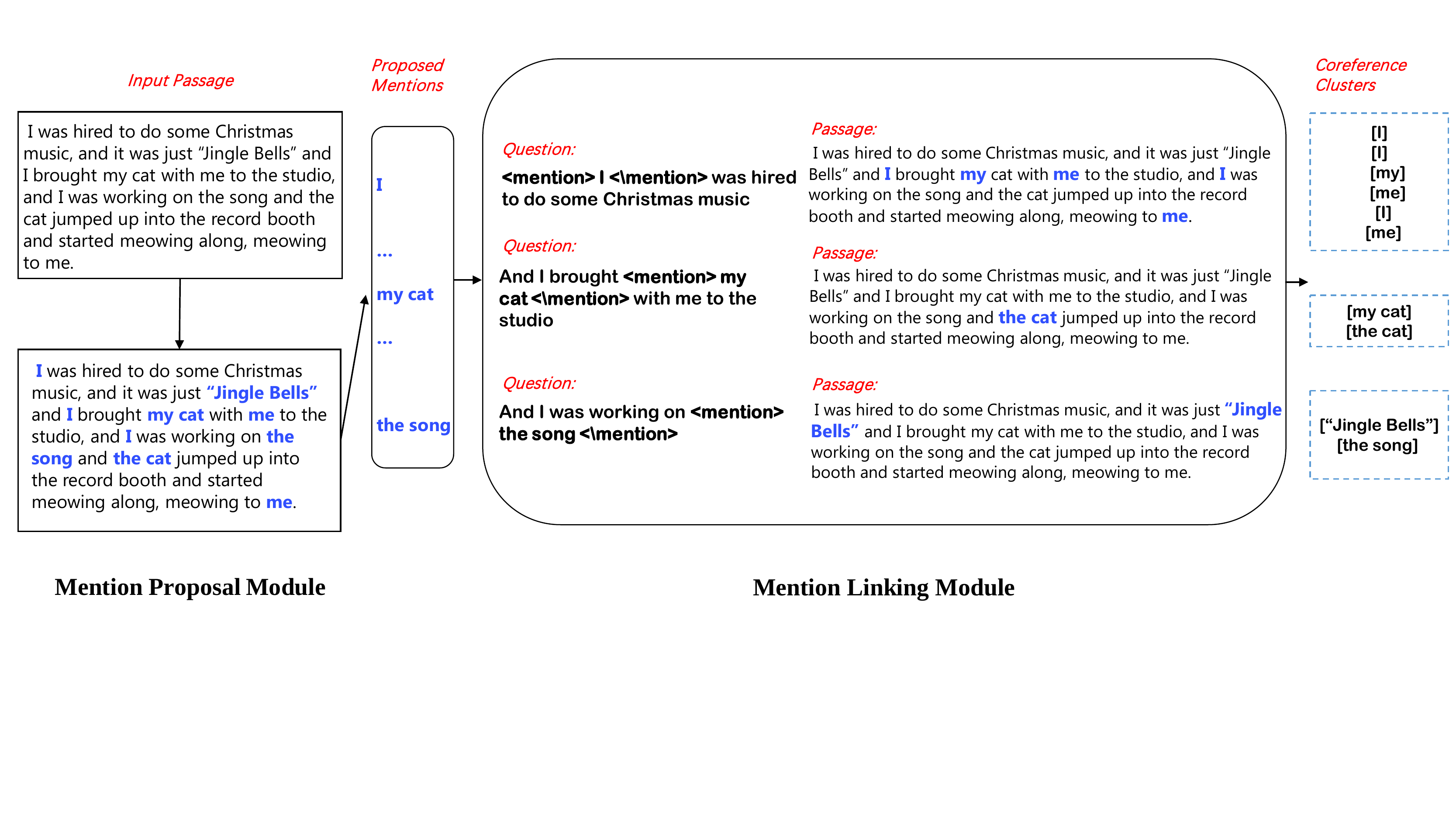}
\caption{The overall architecture of our CorefQA model. The input passage is first fed into the Mention Proposal Module \ref{ssec:mention_pruning} to obtain candidate mentions. Then the Mention Linking Module \ref{ssec:question_answering} is used to extract coreferent mentions from the passage for each proposed mention. The coreference clusters are obtained using the scores produced in the above two stages.}
\label{fig:architecture}

\end{figure*}

\subsection{Formalizing NLP Tasks as question answering}
Machine reading comprehension is a general and extensible task form. Many tasks in natural language processing can be framed as reading comprehension while abstracting away the task-specific modeling constraints. 

\citet{DBLP:journals/corr/abs-1806-08730} introduced the decaNLP challenge, which converts a set of 10 core tasks in NLP to reading comprehension. \citet{DBLP:conf/emnlp/HeLZ15} showed that semantic role labeling annotations could be solicited by using question-answer pairs to represent the predicate-argument structure. \citet{DBLP:conf/conll/LevySCZ17} reduced relation extraction to answering simple reading comprehension questions, yielding models that generalize better in the zero-shot setting. \citet{li2019unified, DBLP:conf/acl/LiYSLYCZL19} cast the tasks of named entity extraction and relation extraction as a reading comprehension problem. In parallel to our work, \citet{DBLP:journals/corr/abs-1908-11141} converted coreference and ellipsis resolution in a question answering format, and showed the benefits of training joint models for these tasks. Their models are built under the assumption that gold mentions are provided at inference time, whereas our model does not need that assumption -- it jointly trains the mention proposal model and the coreference resolution model in an end-to-end manner.
\citet{chada2019gendered} 
 proposed an extractive QA model for the resolution of ambiguous pronouns, 
 and showed better results on the GAP dataset by only  fine-tuning  the  pre-trained BERT model.

\subsection{Data Augmentation}
Data augmentation is a strategy that enables practitioners to significantly increase the diversity of data available for training models. Data augmentation techniques have been explored in various fields such as question answering \citep{DBLP:conf/acl/TalmorB19}, text classification \citep{DBLP:conf/naacl/Kobayashi18} and dialogue language understanding \citep{DBLP:conf/coling/HouLCL18}. In coreference resolution, \newcite{DBLP:conf/naacl/ZhaoWYOC18, DBLP:conf/acl/EmamiTTSSC19, DBLP:conf/naacl/ZhaoWYCOC19} focused on debiasing the gender bias problem; \citet{DBLP:journals/corr/abs-1908-11141} explored the effectiveness of joint modeling of ellipsis and coreference resolution. To the best of our knowledge, we are the first to use existing question answering datasets as data augmentation for coreference resolution.

\section{Model}
In this section, we describe our CorefQA model in detail. The overall architecture is illustrated in Figure \ref{fig:architecture}.

\subsection{Notations}
Given a sequence of input tokens $X = \{x_1, x_2,..., x_n\}$ in a document, where $n$ denotes the length of the document.
 $N = n*(n+1)/2$ denotes the number of all possible text spans in $X$. Let $e_i$ denotes the $i$-th span representation $1\leq i\leq N$, with the start index \textsc{first}(i) and the end index \textsc{last}(i). 
$e_i = \{ x_{\textsc{first}(i)}, x_{\textsc{first}(i)+1},..., x_{\textsc{last}(i)-1}, x_{\textsc{last}(i)}\} $. 
The task of coreference resolution is to determine the antecedents for all possible spans. 
If a candidate span $e_i$ does not represent an entity mention or is not coreferent with any other mentions, a dummy token $\epsilon$ is assigned as its antecedent. The linking between all possible spans $e$ defines the final clustering.

\subsection{Input Representations}
\label{ssec:token_representation}

We use the SpanBERT model \footnote{\url{https://github.com/facebookresearch/SpanBERT}} to obtain input representations following \citet{DBLP:journals/corr/abs-1907-10529}. Each token $x_i$ is associated with a SpanBERT representation $\V{x}_i$. 
Since the speaker information is indispensable for coreference resolution, previous methods \cite{DBLP:conf/naacl/WisemanRS16, DBLP:conf/emnlp/LeeHLZ17, DBLP:journals/corr/abs-1907-10529} usually convert the speaker information into binary features indicating whether two mentions are from the same speaker. However, we use a straightforward strategy that directly concatenates the speaker's name with the corresponding utterance.
This strategy is inspired by recent research in personalized dialogue modeling that use persona information to represent speakers \citep{li2016persona,zhang2018personalizing,mazare2018training}. 
In subsection \ref{ssec:speaker}, we will empirically demonstrate its superiority over the feature-based method in \citet{DBLP:conf/emnlp/LeeHLZ17}. 

To fit long documents into SpanBERT, we use a sliding-window approach that creates a $T$-sized segment after every $T$/2 tokens. Segments are then passed to the SpanBERT encoder independently. The final token representations are derived by taking the token representations with maximum context.

\subsection{Mention Proposal}
\label{ssec:mention_pruning}

Similar to \citet{DBLP:conf/emnlp/LeeHLZ17}, our model considers all spans up to a maximum length $L$ as potential mentions. To improve computational efficiency, we further prune the candidate spans greedily during both training and evaluation. To do so, the mention score of each candidate span consists of three parts: 
(1)  $\V{x}_{\textsc{first}(i)}$ is the start of a span;
(2)  $\V{x}_{\textsc{last}(i)}$ is the end of a span; and 
(3) $\V{x}_{\textsc{first}(i)}$  and $\V{x}_{\textsc{last}(i)}$ form a valid span. 
The third part (i.e., (3)) is necessary because each sentence can contain multiple spans. 
The first part is computed by feeding $\V{x}_{\textsc{first}(i)}$ into a feed-forward layer:
\begin{equation}
\mscore{\V{x}_{\textsc{first}(i)}}= \ffnn{}{[ \V{x}_{\textsc{first}(i)}]}
\label{mention}
\end{equation}
Similarly,
the first part is computed by feeding $\V{x}_{\textsc{last}(i)}$ into a feed-forward layer:
\begin{equation}
\mscore{\V{x}_{\textsc{last}(i)}}= \ffnn{}{[ \V{x}_{\textsc{last}(i)}]}
\label{mention}
\end{equation}
The third part
 computed by feeding the
 concatenation of  $\V{x}_{\textsc{first}(i)}$ and $\V{x}_{\textsc{last}(i)}$ into 
  into a feed-forward layer:
\begin{equation}
\mscore{\V{x}_{\textsc{first}(i)}, \V{x}_{\textsc{last}(i)}}= \ffnn{m}{[ \V{x}_{\textsc{first}(i)}, \V{x}_{\textsc{last}(i)}]}
\label{mention}
\end{equation}
$\ffnn{}$ denotes the feed-forward neural network that computes a nonlinear mapping from the input vector to the mention score. 
The three involved $\ffnn{}$ use separate sets of parameters. 
The overall score for span $i$ being a mention is the average of the three parts:
\begin{equation}
\begin{aligned}
s_m(i) = &[ \mscore{\V{x}_{\textsc{first}(i)}} + \mscore{\V{x}_{\textsc{end}(i)}} \\ 
+ &\mscore{\V{x}_{\textsc{first}(i)}, \V{x}_{\textsc{last}(i)}} ] /3
\end{aligned}
\end{equation}
We only keep up to $\lambda n$ (where $n$ is the document length) spans with the highest mention scores.
\paragraph{Mention Proposal Pretraining}
It is crucial that the mention proposal model is pretrained. Otherwise, most of the proposed mentions that are fed to the linking stage are invalid mentions. 
The mention proposal model is pretrained by jointly training three binary classification models: (1) whether 
$\V{x}_{\textsc{first}(i)}$ is the start of a span; 
(2) whether  $\V{x}_{\textsc{last}(i)}$ is the end of a span; and 
(3) whether $\V{x}_{\textsc{first}(i)}$  and $\V{x}_{\textsc{last}(i)}$  should be combined. 
This leads to the objective of the mention proposal model as follows:
\begin{equation}
\begin{aligned}
\text{Loss(m)} = &\text{sigmoid}(  \mscore{\V{x}_{\textsc{first}(i)}})  \\
+& \text{sigmoid}(  \mscore{\V{x}_{\textsc{end}(i)}}) \\
+ &   \text{sigmoid}( \mscore{\V{x}_{\textsc{first}(i)}, \V{x}_{\textsc{last}(i)}})
\end{aligned}
\end{equation}

\subsection{Mention Linking as Span Prediction}
\label{ssec:question_answering}
Given a mention $e_i$ proposed by the mention proposal network, the role of the mention linking network is to give a score $s_a(i,j)$ for any text span $e_j$, indicating whether $e_i$ and $e_j$ are coreferent. 
We propose to use the question answering framework as the backbone to compute $s_a(i,j)$. It operates on the triplet \{context (X), query (q), answers (a)\}.
The {\bf context} $X$ is the input document.
The {\bf query} $q(e_i)$ is constructed as follows: given $e_i$, we use the sentence that $e_i$ resides in as the query, with the minor modification that we encapsulates $e_i$ with special tokens $<mention></mention>$ .
The {\bf answers} $a$ are the coreferent mentions of $e_i$. 
A query $i$ is considered unanswerable in the following scenarios: (1) the candidate span $e_i$ does not represent an entity mention or (2) the candidate span $e_i$ represents an entity mention but is not coreferent with any other mentions in $X$.

Following \citet{DBLP:conf/naacl/DevlinCLT19}, we represent the input query and the context as a single packed sequence. 
The  for any
span $j=[{\textsc{first}(j)}, ..., {\textsc{last}(j)}]$,
we first compute the score of $i$ being the answer for query $q(e_i)$, denoted by $s_a(j |i)$. 
Let $\V{x}_{\textsc{first}(j)}|i$ and $\V{x}_{\textsc{last}(j)}|i$  respectively denote the representations
for \textsc{first}(j) and \textsc{last}(j) from BERT, where $q(e_i)$ is used as query concatenated to the context. 
$s_a(j |i)$
is computed by feeding the first and the last of its constituent token representations (i.e., $\V{x}_{\textsc{first}(j)}|i$ and $\V{x}_{\textsc{last}(j)}|i$ ) into a feed-forward layer:
\begin{equation}
s_a(j |i) = \textsc{ffnn}_{j|i}{[ \V{x}_{\textsc{first}(j)|i}, \V{x}_{\textsc{last}(j) | i}]}
\label{forward}
\end{equation}
$\textsc{ffnn}_{j|i}$ denotes the feed-forward neural network that computes a nonlinear mapping from the input vector to the mention score. 
Comparing Eq.\ref{forward} with Eq.\ref{mention}, we can observe their  relatedness and  difference: 
both of the equations compute scores for a span. But for Eq.\ref{forward}, the query $q(e_i)$ is additionally used to check whether span $j$ is the answer for $q(e_i)$.

A closer look at Eq.\ref{forward} reveals that it only models the uni-directional coreference relation from $e_i$ to $e_j$, i.e., $e_j$ is the answer for query $q(e_i)$.
This is suboptimal since if $e_i$ is a coreference mention of $e_j$, then $e_j$ should also be the coreference mention $e_i$. 
We thus need to optimize the bi-directional relation between $e_i$ and $e_j$.\footnote{This bidirectional relationship is actually referred to as mutual dependency and has shown to benefit a wide range of NLP tasks such as machine translation \cite{hassan2018achieving} or dialogue generation \cite{li2015diversity}.}
The final score $s_a(i,j)$ is thus given as follows:
\begin{equation}
s_a(i,j) = \frac{1}{2}( s_a(j | i) + s_a(i| j))
\label{pair}
\end{equation}
$s_a(i| j) $ can be computed in the same way as $ s_a(j| i) $, in which $q(e_i)$ is used as the query:
\begin{equation}
s_a(i |j) = \textsc{ffnn}_{i|j}{[ \V{x}_{\textsc{first}(i)|j}, \V{x}_{\textsc{last}(i)|j}]}
\label{forward}
\end{equation}
where $\V{x}_{\textsc{first}(i)}|j$ and $\V{x}_{\textsc{last}(i)}|j$  respectively denote the representations
for \textsc{first}(i) and \textsc{last}(i) from BERT, where $q(e_j)$ is used as query concatenated to the context.

For a pair of text span $e_i$ and $e_j$, the premises for them being coreferent mentions are (1) they are mentions and (2) they are coreferent. 
This makes the overall score $s(i,j)$ for $e_i$ and $e_j$ the combination of Eq.\ref{mention} and Eq.\ref{pair}: 
\begin{equation}
s(i,j) = \lambda [\mscore{i} + \mscore{j}]+  (1-\lambda) s_a (i, j)
\label{global-score}
\end{equation}
$\lambda$ is the hyperparameter to control 
the tradeoff between mention proposal and 
mention linking. 

\subsection{Antecedent Pruning}
Given a document $X$ with length $n$ and the number of spans $O(n^2)$, the computation of Eq.\ref{global-score} for all mention pairs is intractable with the complexity of $O(n^4)$.
Given an extracted mention $e_i$, the computation of Eq.\ref{global-score} for $(e_i, e_j)$ regarding all $e_j$ is still extremely intensive since the computation of the backward span prediction score $s_a( i | j)$ requires running question answering models on all query $q(e_j)$. 
A further pruning procedure is thus needed: For each query $q(e_i)$, we collect $C$ span candidates only based on the $s_a(j|i)$ scores, and then use Eq. \ref{global-score}
to compute the overall scores. 
\subsection{Training}

For each mention $e_i$ proposed by the mention proposal network, it is associated with $C$ potential spans proposed by the mention linking network based on $s(j|i)$, 
we aim to optimize the marginal log-likelihood of all correct antecedents implied by the gold clustering.
Following \citet{DBLP:conf/emnlp/LeeHLZ17}, we append a dummy token $\epsilon$ to the $C$ candidates. The model will output it if none of the $C$ span candidates is coreferent with $e_i$. 
 For each mention $e_i$, the model learns a distribution $P(\cdot)$ over all possible antecedent spans $e_j$ based on the global score $s(i,j)$ from Eq. \ref{global-score}:
\begin{equation}
P(e_j) = \frac{e^{s(i, j)}}{\sum_{j'\in C} e^{s(i, j')}}
 \label{train}
\end{equation}
The mention proposal module and the mention linking module are jointly trained in an end-to-end fashion using training signals from Eq.\ref{train}, with the
SpanBERT parameters shared.

\subsection{Inference}
\label{ssec:mention_clustering}
Given an input document, we can obtain an undirected graph using the overall score, each node of which represents a candidate mention from either the mention proposal module or the mention linking module. 
We prune the graph by keeping the edge whose weight is the largest for each node based on Eq.\ref{train}. Nodes whose closest neighbor is the dummy token $\epsilon$ are abandoned. Therefore, the mention clusters can be decoded from the graph.
 
\subsection{Data Augmentation using Question Answering Datasets}
We hypothesize that the reasoning (such as synonymy, world knowledge, syntactic variation, and multiple sentence reasoning) required to answer the questions are also indispensable for coreference resolution. 
Annotated question answering datasets are usually significantly larger than the coreference datasets due to the high linguistic expertise required for the latter. 
Under the proposed QA formulation, coreference resolution has the same format as the
 existing question answering datasets \citep{DBLP:conf/emnlp/RajpurkarZLL16, DBLP:conf/acl/RajpurkarJL18, DBLP:journals/corr/abs-1908-05803}. In this way, they can readily be used for data augmentation. We thus propose to pre-train the mention linking network on the Quoref dataset \cite{dasigi2019quoref}, 
and the SQuAD dataset \cite{rajpurkar2016squad}. 

\subsection{Summary and Discussion}
Comparing with existing models \cite{DBLP:conf/emnlp/LeeHLZ17, DBLP:conf/naacl/LeeHZ18, DBLP:journals/corr/abs-1908-09091}, the proposed question answering formalization has the flexibility of retrieving mentions left out at the mention proposal stage. However, since we still have the mention proposal model, we need to know in which situation missed mentions could be retrieved and in which situation they cannot.
We use the example in Figure \ref{fig:intro-example} as an illustration, in which \{\textcolor{red}{many people}, \textcolor{red}{They}, \textcolor{red}{themselves}\} are coreferent mentions: If partial mentions are missed by the mention proposal model, e.g., \textcolor{red}{many people} and \textcolor{red}{They}, they can still be retrieved in the mention linking stage when the not-missed mention (i.e., \textcolor{red}{themselves}) is used as query. But, if all the mentions within the cluster are missed, none of them can be used for query construction, which means they all will be irreversibly left out.
Given the fact that the proposal mention network proposes a significant number of mentions, the chance that mentions within a mention cluster are all missed is relatively low (which exponentially decreases as the number of entities increases).
This explains the superiority (though far from perfect) of the proposed model. However, how to completely remove the mention proposal network remains a problem in the field of coreference resolution.

\section{Experiments}

\subsection{Implementation Details}
The special tokens used to denote the speaker's name ($<speaker></speaker>$) and the special tokens used to denote the queried mentions ($<mention></mention>$) are initialized by randomly taking the unused tokens from the SpanBERT vocabulary. The sliding window size $T$ = 512, and the mention keep ratio $\lambda$ = 0.2. The maximum length $L$ for mention proposal = 10 and the maximum number of antecedents kept for each mention $C$ = 50. The SpanBERT parameters are updated by the Adam optimizer \citep{DBLP:journals/corr/KingmaB14} with initial learning rate $1 \times 10^{-5}$ and the task parameters are updated by the Range optimizer \footnote{\url{https://github.com/lessw2020/Ranger-Deep-Learning-Optimizer}} with initial learning rate $2 \times 10^{-4}$.

\subsection{Baselines}
We compare the CorefQA model with previous neural models that are trained end-to-end:
\begin{itemize}
\item e2e-coref \citep{DBLP:conf/emnlp/LeeHLZ17} is the first end-to-end coreference system that learns which spans are entity mentions and how to best cluster them jointly. Their token representations are built upon the GLoVe \citep{DBLP:conf/emnlp/PenningtonSM14} and Turian \citep{DBLP:conf/acl/TurianRB10} embeddings.
\item c2f-coref + ELMo \citep{DBLP:conf/naacl/LeeHZ18} extends \citet{DBLP:conf/emnlp/LeeHLZ17} by combining a coarse-to-fine pruning with a higher-order inference mechanism. Their representations are built upon ELMo embeddings \citep{DBLP:conf/naacl/PetersNIGCLZ18}.
\item c2f-coref + BERT-large\citep{DBLP:journals/corr/abs-1908-09091} builds the c2f-coref system on top of BERT \citep{DBLP:conf/naacl/DevlinCLT19} token representations.
\item EE + BERT-large \citep{DBLP:conf/acl/KantorG19} represents each mention in a cluster via an approximation of the sum of all mentions in the cluster.
\item c2f-coref + SpanBERT-large \citep{DBLP:journals/corr/abs-1907-10529} focuses on pre-training span representations to better represent and predict spans of text.
\end{itemize}

\begin{table*}[t]
\footnotesize
\centering
\setlength{\tabcolsep}{4pt}
\begin{tabular}{lccc@{\hspace{0.5cm}}ccc@{\hspace{0.5cm}}ccc@{\hspace{0.5cm}}c}
\toprule
 & \multicolumn{3}{c}{MUC}& \multicolumn{3}{c}{$\text{B}^3$}& \multicolumn{3}{c}{$\text{CEAF}_{\phi_4}$} \\
 & P & R & F1 & P & R & F1 & P & R & F1 & Avg. F1 \\
\midrule
e2e-coref\citep{DBLP:conf/emnlp/LeeHLZ17} & 78.4 & 73.4 & 75.8 & 68.6 & 61.8 & 65.0 & 62.7& 59.0 &60.8 &67.2\\
c2f-coref + ELMo \citep{DBLP:conf/naacl/LeeHZ18}  & 81.4 & 79.5 & 80.4 & 72.2 & 69.5 & 70.8 & 68.2 & 67.1 & 67.6 & 73.0 \\
EE + BERT-large \citep{DBLP:conf/acl/KantorG19} & 82.6 & 84.1 & 83.4 & 73.3 & 76.2 & 74.7 & 72.4 & 71.1 & 71.8 & 76.6 \\
c2f-coref + BERT-large \citep{DBLP:journals/corr/abs-1908-09091}& 84.7 & 82.4 & 83.5 & 76.5 & 74.0 & 75.3 & 74.1 & 69.8 & 71.9 & 76.9\\
c2f-coref + SpanBERT-large \citep{DBLP:journals/corr/abs-1907-10529}& 85.8 & 84.8 & 85.3 & 78.3 & 77.9 & 78.1 & 76.4& 74.2 & 75.3 & 79.6\\
\midrule
CorefQA + SpanBERT-base & 85.2 & 87.4 & 86.3 & 78.7 & 76.5 & 77.6 & 76.0 & 75.6 & 75.8 & 79.9 (+0.3)\\
CorefQA + SpanBERT-large & \textbf{88.6} & \textbf{87.4} & \textbf{88.0} &  \textbf{82.4} & \textbf{82.0} & \textbf{82.2} & \textbf{79.9} & \textbf{78.3} & \textbf{79.1} & \textbf{83.1} (+3.5)\\
\bottomrule
\end{tabular}
\caption{Evaluation results on the English CoNLL-2012 shared task. The average F1 of  MUC, $\text{B}^3$, and $\text{CEAF}_{\phi_4}$ is the main evaluation metric. Ensemble models are not included in the table for a fair comparison.}
\label{tab:ontonotes}
\end{table*}

\begin{table}[t]
\footnotesize
\centering
\setlength{\tabcolsep}{4pt}
\begin{tabular}{lcccc}
\toprule
Model & \textbf{M} & \textbf{F} & \textbf{B} & \textbf{O} \\
\midrule
e2e-coref & 67.2 & 62.2 & 0.92 & 64.7 \\
c2f-coref + ELMo& 75.8 & 71.1 & 0.94 & 73.5 \\
c2f-coref + BERT-large & 86.9 & 83.0 & 0.95 & 85.0 \\
\midrule
c2f-coref + SpanBERT-large & 88.8 & 84.9 & 0.96 & 86.8 \\
CorefQA + SpanBERT-large   & \textbf{88.9} & \textbf{86.1} & \textbf{0.97} & \textbf{87.5} \\
\bottomrule
\end{tabular}
\caption{CorefQA achieves the state-of-the-art performance on all metrics including F1 scores on \textbf{M}asculine and \textbf{F}eminine examples, a \textbf{B}ias factor (F / M) and the \textbf{O}verall F1 score.}
\label{tab:gap_res}
\end{table}

\subsection{Results on CoNLL-2012 Shared Task}
The English data of CoNLL-2012 shared task \citep{DBLP:conf/conll/PradhanMXUZ12} contains 2,802/343/348 train/development/test documents in 7 different genres. The main evaluation is the average of three metrics -- MUC \citep{DBLP:conf/muc/VilainBACH95}, $\text{B}^3$ \citep{Bagga98algorithmsfor}, and $\text{CEAF}_{\phi_4}$ \citep{DBLP:conf/naacl/Luo05} on the test set according to the official CoNLL-2012 evaluation scripts \footnote{\url{http://conll.cemantix.org/2012/software.html}}.

We compare the CorefQA model with several baseline models in Table \ref{tab:ontonotes}. Our CorefQA system achieves a huge performance boost over existing systems: With SpanBERT-base, it achieves an F1 score of 79.9, which already outperforms the previous SOTA model using SpanBERT-large by 0.3. With SpanBERT-large, it achieves an F1 score of 83.1, with a 3.5 performance boost over the previous SOTA system.

\subsection{Results on GAP}
The GAP dataset \citep{DBLP:journals/tacl/WebsterRAB18} is a gender-balanced dataset that targets the challenges of resolving naturally occurring ambiguous pronouns. It comprises 8,908 coreference-labeled pairs of (ambiguous pronoun, antecedent name) sampled from Wikipedia.

We follow the protocols in \citet{DBLP:journals/tacl/WebsterRAB18, 
DBLP:journals/corr/abs-1908-09091} and use the off-the-shelf resolver trained on the CoNLL-2012 dataset to get the performance of the GAP dataset. Table \ref{tab:gap_res} presents the results. We can see that the proposed CorefQA model achieves state-of-the-art performance on all metrics on the GAP dataset.

\section{Ablation Study and Analysis}
\begin{table}[t!]
\newcolumntype{Y}{>{\centering\arraybackslash}X}
\newcommand{\colindent}{\;}
\setlength{\tabcolsep}{0.25em}
\centering
\begin{tabularx}{\linewidth}{l Y Y}
\toprule
 & Avg.~F1 & $\Delta$ \\
\midrule
CorefQA & 83.4 &  \\
$-\text{--}$ SpanBERT & 79.6 & -3.8\\
$-\text{--}$ Mention Proposal Pre-training & 75.9 & -7.5\\
$-\text{--}$ Question Answering & 75.0 & -8.4\\
$-\text{--}$ Quoref Pre-training & 82.7 & -0.7\\
$-\text{--}$ Squad Pre-training & 83.1 & -0.3\\
\bottomrule
\end{tabularx}

\caption{Ablation studies on the CoNLL-2012 development set. SpanBERT token representations, the mention-proposal pre-training, and the question answering pre-training all contribute significantly to the good performance of the full model.}
\label{tab:ablations}
\end{table}
We perform comprehensive ablation studies and analyses on the CoNLL-2012 development dataset. Results are shown in Table \ref{tab:ablations}.
\subsection{Effects of Different Modules in the Proposed Framework}
\paragraph{Effect of SpanBERT} 
Replacing SpanBERT with vanilla BERT leads to a 3.5 F1 degradation. This verifies the importance of span-level pre-training for coreference resolution and is consistent with previous findings \citep{DBLP:journals/corr/abs-1907-10529}. 

\paragraph{Effect of Pre-training Mention Proposal Network} 
Skipping the pre-training of the mention proposal network using golden mentions results in a 7.2 F1 degradation, which is in line with our expectation. 
A randomly initialized mention proposal model implies that mentions are randomly selected. Randomly selected mentions will mostly be transformed to unanswerable queries. This makes it hard for the question answering model to learn at the initial training stage, leading to inferior performance.

\paragraph{Effect of QA pre-training on the augmented datasets} 
One of the most valuable strengths of converting anaphora resolution to question answering is that existing QA datasets can be readily used for data augmentation purposes. We see a contribution of 0.7 F1 from pre-training on the Quoref dataset \citep{DBLP:journals/corr/abs-1908-05803} and a contribution of 0.3 F1 from pre-training on the SQuAD dataset \citep{DBLP:conf/emnlp/RajpurkarZLL16}.

\paragraph{Effect of Question Answering} We aim to study the pure performance gain of the paradigm shift from mention-pair scoring to query-based span prediction. For this purpose, we replace the mention linking module with the mention-pair scoring module described in \citet{DBLP:conf/naacl/LeeHZ18}, while others remain unchanged. We observe an 8.1 F1 degradation in performance, demonstrating the significant superiority of the proposed question answering framework over the mention-pair scoring framework. 

\subsection{Analyses on speaker modeling strategies}
\label{ssec:speaker}
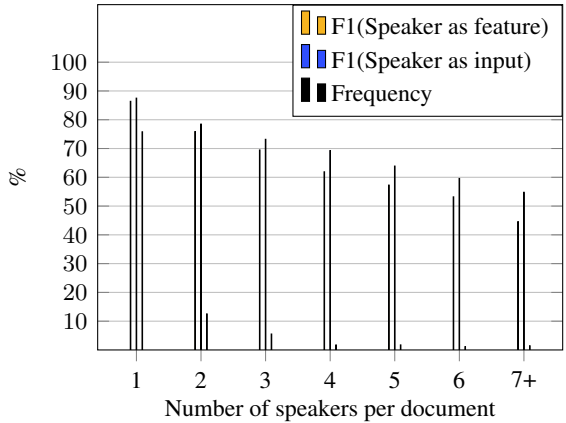
\begin{figure}[t!]
    \centering
        \begin{tikzpicture}[trim left=-0.6cm]
        \begin{axis}[
    	width=\columnwidth,
	    height=0.8\columnwidth,
	    legend cell align=left,
	    legend style={at={(1, 1)},anchor=north east,font=\small},
        xtick={1, 2, 3, 4, 5, 6, 7},
        xticklabels={1, 2, 3, 4, 5, 6, 7+},
   	 	ytick={10, 20, 30, 40, 50, 60, 70, 80, 90, 100},
   		ymin=0, ymax=120,
	    xtick pos=left,
        ybar,
    	ymajorgrids=true,
		font=\small,
        bar width=0.2,
    	xlabel=Number of speakers per document,
        ylabel=\%,
    	ylabel style={yshift=-1ex,},
        ]
    \addplot[fill=head] coordinates {
(1,86.4)
(2,75.9)
(3,69.5)
(4,61.9)
(5,57.3)
(6,53.2)
(7,44.6)
            };
    \addlegendentry{F1(Speaker as feature)}
            
    \addplot[fill=shannon] coordinates {
(1,87.5)
(2,78.5)
(3,73.2)
(4,69.3)
(5,63.9)
(6,59.6)
(7,54.8)

            };
    \addlegendentry{F1(Speaker as input)}
    
    \addplot[fill=freq] coordinates {
 (1, 75.80174927113703)
 (2, 12.536443148688047)
 (3, 5.539358600583091)
 (4, 1.749271137026239)
 (5, 1.749271137026239)
 (6, 1.1661807580174928)
 (7, 1.457725947521866)
            };
    \addlegendentry{Frequency}
    
    \end{axis}
    \end{tikzpicture}
    \caption{Performance on the development set of the CoNLL-2012 dataset with various number of speakers. F1(Speaker as feature): F1 score for the strategy that treats speaker information as a mention-pair feature. F1(Speaker as input): F1 score for our strategy that treats speaker names as token input. Frequency: percentage of documents with specific number of speakers.}
    \label{fig:analyze_speaker}
\end{figure}
We compare our speaker modeling strategy (denoted by {\it Speaker as input}), which directly concatenates the speaker's name with the corresponding utterance, with the strategy in 
 \newcite{DBLP:conf/naacl/WisemanRS16, DBLP:conf/emnlp/LeeHLZ17, DBLP:journals/corr/abs-1907-10529} (denoted by {\it Speaker as feature}), which converts speaker information into binary features indicating whether two mentions are from the same speaker.
We show the average F1 scores breakdown by documents according to the number of their constituent speakers in Figure \ref{fig:analyze_speaker}. 

Results show that the proposed strategy performs significantly better on documents with a larger number of speakers. Compared with the coarse modeling of whether two utterances are from the same speaker, a speaker's name can be thought of as speaker ID in persona dialogue learning \cite{li2016persona,zhang2018personalizing,mazare2018training}. Representations learned for names have the potential to better generalize the global information of the speakers in the multi-party dialogue situation, leading to better context modeling and thus better results.

\subsection{Analysis on the Overall Mention Recall}
\begin{figure}[t!]
\vspace{-7pt}
\begin{tikzpicture}
\begin{axis}[
    	width=1.0\columnwidth,
	    height=0.8\columnwidth,
	    legend cell align=left,
	    legend style={at={(1, 0)},anchor=south east,font=\small},
	    xtick={10, 20, 30, 40, 50},
   	 	ytick={10, 20, 30, 40, 50, 60, 70, 80, 90, 100},
   		ymin=30, ymax=100,
   		xtick pos=left,
   		xtick align=outside,
	    xmin=5,xmax=55,
	    mark options={mark size=2},
		font=\small,
   	 	ymajorgrids=true,
    	xmajorgrids=true,
    	xlabel=the number of spans $\lambda$ kept per word,
        ylabel=Mention Recall (\%),
    	ylabel style={yshift=-1ex,}]
    	
\addplot[
    color=red,
    line width=1.5pt
    ]
    coordinates {
(10, 56.81)(15, 74.12)(20, 84.89)(25, 88.19)(30,90.90)(40,92.57)(50, 93.28)
    };
    \addlegendentry{\citet{DBLP:journals/corr/abs-1907-10529} (various $\lambda$)}
    
\addplot[
    only marks,
    color=red,
    mark=*,
    mark size=10pt
    ]
    coordinates {
(40,92.67)
    };
    \addlegendentry{\citet{DBLP:journals/corr/abs-1907-10529}  (actual $\lambda$)}

\addplot[
    color=shannon,
    line width=1.5pt
    ]
    coordinates {
(10, 73.71)(15, 87.62)(20, 92.79)(25, 93.27)(30,94.36)(40,95.06)(50, 96.19)
    };
    \addlegendentry{Our model (various $\lambda$)}
    
\addplot[
    only marks,
    color=shannon,
    mark=square*,
    mark size=10pt
    ]
    coordinates {
    (20, 92.79)
    };
    \addlegendentry{Our model (actual $\lambda$)}
\end{axis}
\end{tikzpicture}
\caption{Change of mention recalls as we increase the number of spans $\lambda$ kept per word.}
\label{fig:recall}

\end{figure}
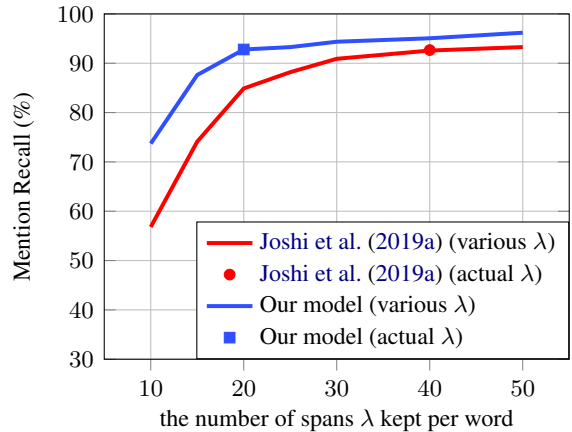
Since the proposed framework has the potential to retrieve 
mentions missed at the mention proposal stage, we expect it to have higher overall mention recall rate than previous models
\citep{DBLP:conf/emnlp/LeeHLZ17, DBLP:conf/naacl/LeeHZ18, DBLP:conf/acl/ZhangSYXR18, DBLP:conf/acl/KantorG19}. 

We examine the proportion of gold mentions covered in the development set as we increase the hyperparameter $\lambda$ (the number of spans kept per word) in Figure \ref{fig:recall}.
Our model consistently outperforms the baseline model with various values of $\lambda$. Notably, our model is less sensitive to smaller values of $\lambda$. This is because missed mentions can still be retrieved at the mention linking stage. 

\subsection{Qualitative Analysis}
\begin{table}[t!]
\setlength{\tabcolsep}{0.3em}
\centering
\fontsize{10.5}{12.6}\selectfont
\def\tabularxcolumn#1{m{#1}}
\begin{tabularx}{\linewidth}{cX}
\toprule

1 & [\textbf{Freddie Mac}] is giving golden parachutes to two of its ousted executives. \ldots Yesterday Federal Prosecutions announced a criminal probe into [\textbf{the company}]. \\
\midrule
2 & [\textbf{A traveling reporter}] now on leave and joins us to tell [\textbf{her}] story. Thank [\textbf{you}] for coming in to share this with us.  \\
\midrule
3 &  \textit{Paula Zahn:} [\textbf{Thelma Gutierrez}] went inside the forensic laboratory where scientists are trying to solve this mystery.  

\textit{Thelma Gutierrez:} In this laboratory alone [\textbf{I}] 'm surrounded by the remains of at least twenty different service members who are in the process of being identified so that they too can go home. \\ 

\bottomrule
\end{tabularx}
\caption{Example mention clusters that were correctly predicted by our model, but wrongly predicted by c2f-coref + SpanBERT-large. Bold spans in brackets represent coreferent mentions. Italic spans represent the speaker's name of the utterance.}
\label{tab:examples}
\end{table}
We provide qualitative analyses to highlight the strengths of our model in Table \ref{tab:examples}. 

Shown in Example 1, by explicitly formulating the anaphora identification of \textbf{the company} as a query, our model uses more information from a local context, and successfully identifies \textbf{Freddie Mac} as the answer from a longer distance.

The model can also efficiently harness the speaker information in a conversational setting. In Example 3, it would be difficult to identify that [\textbf{Thelma Gutierrez}] is the correct antecedent of mention [\textbf{I}] without knowing that \textit{Thelma Gutierrez} is the speaker of the second utterance. However, our model successfully identifies it by directly feeding the speaker's name at the input level.

\section{Conclusion}
In this paper, we present CorefQA, a coreference resolution model that casts anaphora identification as the task of query-based span prediction in question answering. We showed that the proposed formalization can successfully retrieve mentions left out at the mention proposal stage. It also makes data augmentation using a plethora of existing question answering datasets possible. Furthermore, a new speaker modeling strategy can also boost the performance in dialogue settings. Empirical results on two widely-used coreference datasets demonstrate the effectiveness of our model. In future work, we will explore novel approaches to generate the questions based on each mention, and evaluate the influence of different question generation methods on the coreference resolution task.

\section*{Acknowledgement}
We thank all anonymous reviewers for their comments and suggestions. 
The work is supported by the National Natural Science Foundation of China (NSFC No. 61625107 and 61751209). 

\bibliography{CorefQA}
\bibliographystyle{acl_natbib}

\end{document}